\documentclass[letterpaper, 10 pt, journal, twoside]{IEEEtran}

\usepackage{amsmath,amsfonts}
\usepackage{algorithmic}
\usepackage{algorithm}
\usepackage{array}
\usepackage[caption=false,font=normalsize,labelfont=sf,textfont=sf]{subfig}
\usepackage{textcomp}

\usepackage{stfloats}
\usepackage{url}
\usepackage{verbatim}
\usepackage{graphicx}
\usepackage{cite}
\usepackage{multirow}
\usepackage{booktabs}
\usepackage{makecell}

\usepackage{colortbl}
\usepackage[table]{xcolor} 
\usepackage{float}
\usepackage{ulem}
\usepackage{calligra} 
\ifCLASSINFOpdf
\else
\fi
\begin{document}
%
\title{LIV-GaussMap: LiDAR-Inertial-Visual Fusion for Real-time 3D Radiance Field Map Rendering}

\newif\ifshowpre 
\showprefalse 

\newcommand{\pre}[1]{\ifshowpre\textcolor{blue}{#1}\fi} 

\newcommand{\now}[1]{\textcolor{black}{#1}}

\author{
\thanks{Manuscript received January 17, 2024; Revised April 1, 2024; Accepted April 15, 2024.}
\thanks{This paper was recommended for publication by Editor Pascal Vasseur upon evaluation of the Associate Editor and Reviewers’ comments.}
Sheng Hong$^{1,*}$, Junjie He$^{2,*}$, Xinhu Zheng$^{2}$, Chunran Zheng$^{3}$, Shaojie Shen$^{1\dagger}$
\thanks{* Equal contribution.
$^{\dagger}$ Corresponding Author ({\tt\footnotesize eeshaojie@ust.hk}).
}
\thanks{
$^{1}$ Department of Electronic Computer Engineering, The Hong Kong University of Science and Technology, Hong Kong SAR, China. 
}
\thanks{
$^{2}$ System hub, The Hong Kong University of Science and Technology (Guangzhou), Guangzhou, China.
}
\thanks{
$^{3}$ Department of Mechanical Engineering, The University of Hong Kong, Hong Kong SAR, China. 
}
\thanks{Digital Object Identifier (DOI): see top of this page.}
}


\markboth{IEEE Robotics and Automation Letters. Preprint Version. Accepted April~2024}
{Hong \MakeLowercase{\textit{et al.}}: LIV-GaussMap: LiDAR-Inertial-Visual Fusion for Real-time 3D Radiance Field Map Rendering}

\maketitle

\begin{abstract}

We introduce an integrated precise LiDAR, Inertial, and Visual (LIV) multimodal sensor fused mapping system that builds on the differentiable \pre{surface splatting }\now{Gaussians} to improve the mapping fidelity, quality, and structural accuracy.
Notably, this is also a novel form of tightly coupled map for LiDAR-visual-inertial sensor fusion.

This system leverages the complementary characteristics of LiDAR and visual data to capture the geometric structures of large-scale 3D scenes and restore their visual surface information with high fidelity.
\pre{The initial poses for surface Gaussian scenes are obtained using a LiDAR-inertial system with size-adaptive voxels. Then, we optimized and refined the Gaussians by visual-derived photometric gradients to optimize the quality and density of LiDAR measurements.}
\now{The initialization for the scene's surface Gaussians and the sensor's poses of each frame are obtained using a LiDAR-inertial system with the feature of size-adaptive voxels. Then, we optimized and refined the Gaussians using visual-derived photometric gradients to optimize their quality and density. }

Our method is compatible with various types of LiDAR, including solid-state and mechanical LiDAR, supporting both repetitive and non-repetitive scanning modes. 
Bolstering structure construction through LiDAR and facilitating real-time generation of photorealistic renderings across diverse LIV datasets. 
It showcases notable resilience and versatility in generating real-time photorealistic scenes potentially for digital twins and virtual reality, while also holding potential applicability in real-time SLAM and robotics domains.

We release our software and hardware and self-collected datasets on Github\footnote[1]{https://github.com/sheng00125/LIV-GaussMap} to benefit the community.

\end{abstract}


%
\IEEEpeerreviewmaketitle

\vspace{-0.0cm}
\begin{figure}[ht!]
	\centering
	\includegraphics[width=1\linewidth]{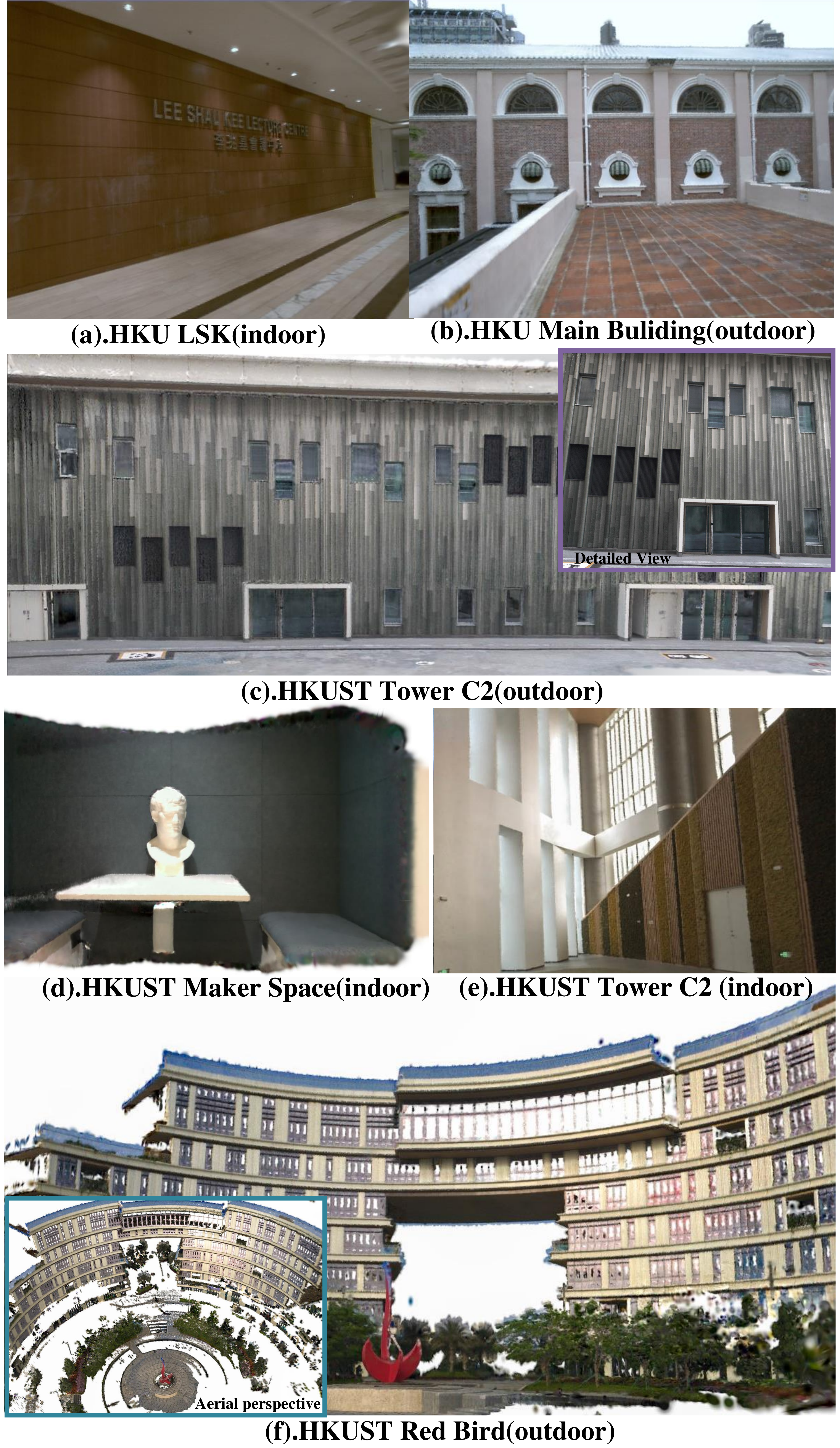}
	\caption{
The real-world experiments were performed in both public datasets and private datasets, including both small-scale indoor environments and large-scale outdoor settings. The image shows our radiance field map of HKU LSK(a), HKU Main Building(b), HKUST GZ Tower C2 outdoor(c) and indoor(e), HKUST GZ Makerspace(d), HKUST GZ Red Bird(f).}

	\vspace{-2pt}
	\label{fig.scheme}
\end{figure}
\vspace{-0.0cm}

\vspace{-0.0cm}
\begin{figure*}[t!]
	\centering
	\includegraphics[width=1.0\linewidth]{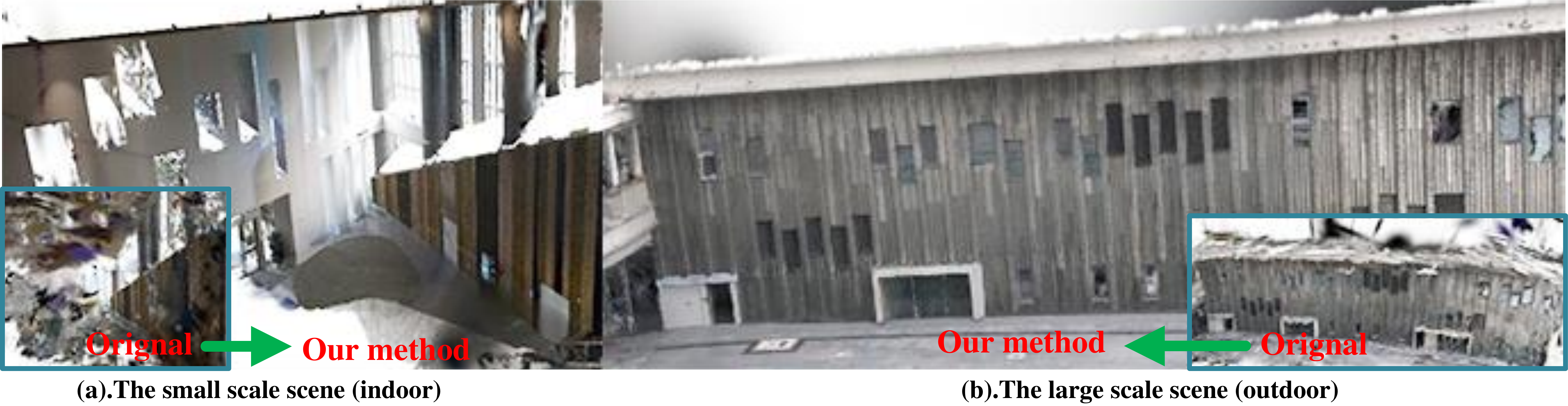}
	\caption{
The figure shows the aerial perspective of the indoor and outdoor scenes of HKUST GZ Tower C1. In contrast to the vision build-up structure \pre{of }\now{by} 3D-GS~\cite{kerbl20233d}, 
our approach yields a more refined structure \pre{without }\now{with few} artifacts.
 }
	\vspace{-2pt}
	\label{fig.structure}
\end{figure*}
\vspace{-0.0cm}

\section{Introduction}
\label{sec:intro}


Simultaneous localization and mapping (SLAM), essential for autonomous navigation, combines map construction of unknown environments with tracking an agent's location~\cite{xu2022point}. Traditional SLAM systems, limited by single sensors, \pre{like }\now{such as} cameras or LiDAR, face challenges \pre{like }\now{such as} light sensitivity or depth perception issues. The multimodal fusion of sensors in SLAM addresses these by integrating data from cameras, LiDAR, and IMU, improving the precision and robustness of the map~\cite{zuo2019lic, zhu2021camvox, shan2021lvi, lin2021r, lin2022r, zheng2022fast}. 
\pre{Key developments in this area include LiDAR-inertial visual odometry (LIVO) and advanced fusion techniques like Kalman filters, with notable systems like LIC-Fusion, Camvox, LVI-SAM, R2LIVE, R3LIVE and FAST-LIVO, significantly improving SLAM's efficiency and precision~\cite{zuo2019lic, zhu2021camvox, shan2021lvi, lin2021r, lin2022r, zheng2022fast}.} 

However, existing LiDAR-inertial visual systems are predominantly designed for scenarios with Lambertian surfaces based on the assumption that the environment exhibits isotropic photometric properties across different viewing directions. Visual information within these systems is typically 
\pre{
represented as image patches associated with 3D points \cite{zheng2022fast} or colored pixel\cite{lin2021r} \cite{lin2022r}.
}
\now{
represented as 3D points associated with image patches~\cite{zheng2022fast} or colored pixels~\cite{lin2022r}. 
}
Tracking and mapping in environments with non-Lambertian surfaces like glass or reflective metal are challenging due to their varying reflective properties. Overcoming this requires specialized sensors or algorithms.
Recent advances in novel view synthesis have shown the ability to generate impressive photorealistic images from new perspectives. These methods employ implicit representations \pre{like }\now{such as} neural radiation fields (NeRF)\cite{mildenhall2021nerf} or explicit representations such as meshes and signed distance functions, including the emerging technique of 3D Gaussian splatting~\cite{kerbl20233d}. By reconstructing the scene's geometric structures while preserving visual integrity with harmonic spherical function, these approaches enable the creation of highly realistic images.

However, in the field of novel view synthesis, the focus on high PSNR often neglects the map structure, leading to poor extrapolation performance, 
\pre{crucial for robotics. Techniques like COLMAP and SfM~\cite{schonberger2016structure} are limited in low-texture scenes. Multimodal sensor fusion improves this, enhancing geometric accuracy and enabling denser, more precise maps.}
\now{which is crucial for robotics. Techniques such as COLMAP and SfM~\cite{schonberger2016structure} are limited in low-texture scenes. Multimodal sensor fusion improves this, enhancing geometric accuracy and enabling more dense and precise maps.}

Overall, the primary contributions of this work can be summarized as follows:

\begin{itemize}
  \item We propose constructing a dense and precise map of the scene by utilizing the \pre{structure for the planar surface in }\now{Gaussians} measurement from the LiDAR-inertial system. This measurement allows us to accurately represent the characteristics of \now{the scene's} surface and create a detailed map.
  \item We propose building up the LiDAR-visual map with differentiable \pre{ellipsoidal }Gaussians with spherical harmonic coefficients, which implies the visual measurement information from different viewing directions. This approach enables real-time rendering with photorealistic performance, enhancing the accuracy and realism of the map.
  \item \pre{We further optimize the structure of the map by incorporating differentiable ellipsoidal Gaussians to mitigate the issue of an unreasonable distribution of point clouds caused by the critical inject angle during scanning, addressing the challenges of unevenly distributed or inaccurately measured point clouds.}
  \now{
  We propose further optimizing the structure of the map by incorporating differentiable ellipsoidal surface Gaussians in order to mitigate the issue of an unreasonable distribution of point clouds caused by the critical inject-angle during scanning, addressing the challenges of unevenly distributed or inaccurately measured point clouds.}
  \item All related software and hardware packages and self-collecting datasets will be publicly available to benefit the community.
\end{itemize}

To our knowledge, this study is the first to utilize multimodal sensor fusion to build a precise and photorealistic Gaussian map. By combining the accurate map from the LiDAR-inertial system with visual photometric measurements, we achieve a comprehensive and detailed representation of the environment.

Our proposed method has undergone rigorous testing and validation on diverse public real-world datasets, including different types of LiDAR \pre{like }\now{, such as} the mechanical Ouster OS1-128, semi-mechanical Livox Avia, and solid-state Realsense L515. The \pre{scene for evaluation }\now{evaluation datasets} covered both indoor (bounded scene) and outdoor (unbounded scene). The experimental results confirm the effectiveness of our algorithm in efficiently capturing and storing image information from multiple viewpoints. This capability enables the rendering of novel views with improved performance.

\begin{figure*}[t]
	\centering
	\includegraphics[width=1.0\linewidth]{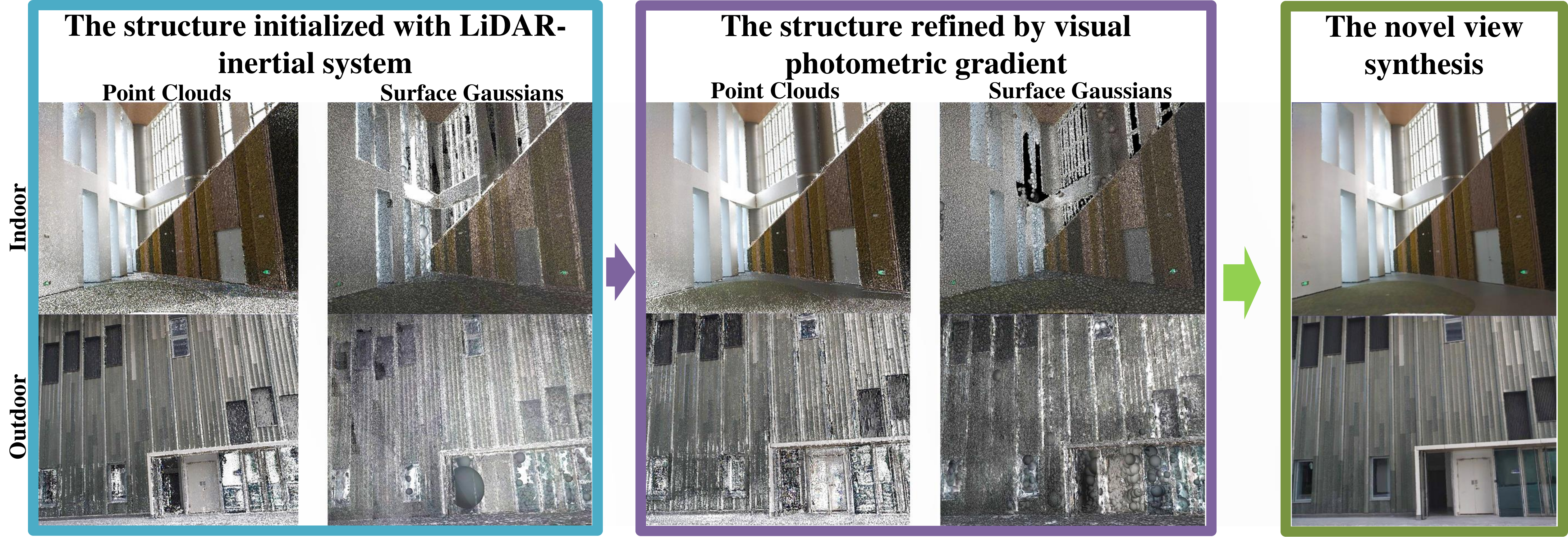}
	\caption{
The construction process of the map is illustrated in the above figure.
\pre{(1). Initially, the Gaussians of the scene are derived from a Kalman-filtered LiDAR-inertial system. The surfaces of 3D objects within the scene are estimated using LiDAR measurements. The Gaussians expand along the surface, resulting in an initial colored point cloud. This further develops into the ellipsoidal surface Gaussians.}
\now{(1). Initially, the Gaussians of the scene are derived from a Kalman-filtered LiDAR-inertial system. The surfaces of the scene are estimated using LiDAR measurements and are further developed into ellipsoidal surface Gaussians.}
\pre{(2). We then enhance the Gaussian distribution utilizing photometric gradients, leading to an optimized point cloud and an optimized Surface Gaussian. This optimized map allows us to synthesize new views with precise photometry and generate maps that are devoid of any gaps.}
\now{(2). We further optimize the Gaussians by using photometric gradients. This optimized map allows us to synthesize new views with precise photometry and generate a hole-free map.}
}
	\vspace{-2pt}
	\label{fig.scheme}
\end{figure*}
\vspace{-0.0cm}

\section{Related works}
\label{sec:related}

\subsection{Related work about Mapping with Multi-modal Sensor}

In the realm of robotics, multi-modal sensor fusion for localization, such as LiDAR-inertial visual odometry(LIVO), is being extensively researched. LiDAR can deliver accurate geometric measurements of real-world environments, while cameras provide detailed 2D imagery of textures and appearances of the environment. Meanwhile, inertial navigation systems provide high-frequency motion measurements. The integration of these sensors is considered ideal for robotic applications.\pre{Numerous works in this area, such as LIC-Fusion~\cite{zuo2019lic}, R2LIVE~\cite{lin2021r}, LVI-SAM~\cite{shan2021lvi}, Camvox~\cite{zhu2021camvox}, R3LIVE~\cite{lin2022r} and FAST-LIVO~\cite{zheng2022fast} have been contributing significantly by enhancing perception capabilities in robotics.}

\pre{
Among them, R3LIVE~\cite{lin2022r} and FAST-LIVO~\cite{zheng2022fast} adopt the tightly coupled iterative error state Kalman filtering methods for multi-modal sensor fusion. They provide accurate and real-time odometry and generate colored point clouds. R3LIVE uses photometric error in RGB-colored point clouds, while FAST-LIVO utilizes warped image patches from diverse viewpoints, similar to SVO. These approaches highlight the flexibility of multi-modal sensor fusion in robotics. LIV-based SLAM systems can generate high-density colored point clouds for realistic visualization. However, these point clouds may have holes and lack photometric realism. Moreover, for anisotropic, non-Lambertian surfaces like glass and metal, the RGB values can vary across different viewpoints, leading to a blurred RGB point cloud map. 
In ~\cite{yuan2022efficient}, an efficient LIO method is presented, employing adaptive voxels with plane features for improved scene mapping and precise LiDAR scan registration.
}
\now{
The trend of multisensor fusion is developing from loose to tightly coupled, exhibiting increased robustness in complex environments.
Notably, frameworks such as Zuo et al.'s LIC-fusion\cite{zuo2019lic} and its successor, LIC-Fusion 2.0, have achieved significant improvements in accuracy and robustness by tightly integrating IMU, visual, and LiDAR data. This integration is facilitated by the novel tracking of plane features, which is a key innovation that contributes to the enhanced performance of the system.
Parallelly, LVI-SAM\cite{shan2021lvi} and R2LIVE\cite{lin2021r} have further advanced the state-of-the-art by tightly combining LiDAR, visual, and inertial data, guaranteeing the robustness of the systems even when encountering sensor malfunctions and in challenging circumstances.
Among them, R3LIVE uses the photometric error of the RGB-colored point clouds for observation, while FAST-LIVO uses the map in the form of points attached with image patches from various viewpoints, similar to SVO~\cite{Forster17troSVO}. These approaches highlight the real-time performance and accuracy of multimodal sensor fusion in robotics, generating colorized point clouds of the scene for visualization.
However, these point clouds are not hole-free and lack photometric realism. Moreover, for anisotropic, non-Lambertian surfaces like glass and metal, the appearance can vary across different viewpoints, leading to blurred colorized point clouds.
}

\pre{
In the context of voxel-based mapping and odometry methods, \cite{lin2022r} and \cite{zheng2022fast} map the world with fixed-size voxels, while \cite{yuan2022efficient} build up the voxel with adaptive size. 
These approaches model the surface in a scene with Gaussian distribution, which resembles surface splatting~\cite{kerbl20233d}~\cite{zhang2022differentiable}~\cite{zwicker2001surface} used for novel view synthesis in computer graphics and 3D visualization.
}
\now{
In the work of~\cite{yuan2022efficient}, an efficient LiDAR-inertial odometry (LIO) method called VoxelMap is proposed. It maps scenes using adaptive voxels that incorporate plane features, such as normal vectors and covariances. This probabilistic representation of the surface allows for the accurate registration of new LiDAR scans.
In the realm of voxel-based mapping and odometry techniques, \cite{lin2022r} and \cite{zheng2022fast} maintain a map with voxels of constant size, while \cite{yuan2022efficient} utilize a method in which the voxel size is dynamically adjusted.
Both these LIO and LIVO represent the structure of the world using an ellipsoidal Gaussian distribution, which resembles surface Gaussians~\cite{kerbl20233d}~\cite{zhang2022differentiable}~\cite{zwicker2001surface} used for novel view synthesis in computer graphics and 3D visualization.
}

\subsection{Related Work about Novel View Synthesis}
\pre{
Novel view synthesis has progressed with continuous radiance field modeling, using explicit representations like meshes, point clouds, SDFs, or implicit ones like NeRF~\cite{mildenhall2021nerf}. These methods, differing from traditional SLAM systems with discrete point clouds, create more photorealistic images by treating scenes as continuous, viewpoint-dependent functions. Instant-NGP~\cite{muller2022instant} further accelerates this with its multiresolution grid structure, enabling real-time rendering and faster training.
}
\now{
The representation of a map can be explicit, such as meshes, point clouds, and signed distance fields (SDF), or implicit, such as neural radiation fields (NeRF)~\cite{mildenhall2021nerf}. Implicit representation for novel view synthesis has seen advances in the modeling of scenes as continuous radiance fields. 
These methods, which differ from traditional SLAM systems modeling scenes with discrete point clouds, create more photorealistic images by treating scenes as continuous, viewpoint-dependent functions.  
Instant-NGP~\cite{muller2022instant} improves this procedure by using a multiresolution hash grid to encode spatial information and spherical harmonics to encode angular information. This reasonable positional encoding solution facilitates a more streamlined neural network, allowing real-time rendering and faster training.}
Mip-NeRF 360~\cite{barron2022mip} addresses unbounded scenes and sampling issues with nonlinear parameterization and new regularizers. Despite implicit representations using neural networks for high-fidelity, low-memory synthesis, they remain computationally intensive. Recent efforts have focused on explicit map representations \pre{like }\now{, such as} using spherical harmonics for voxel-based volumetric density.

PlenOctrees, introduced by Yu et al.\cite{yu2021plenoctrees}, utilize volumetric rendering with spherical harmonics (SHs) to model rays from various directions, offering a compact and efficient way to represent complex 3D scenes. Plenoxels, proposed by Fridovich-Keil et al.\cite{fridovich2022plenoxels}, represent scenes as sparse 3D grids using SHs, optimized through gradient methods without the need for neural networks, significantly reducing computational requirements, and achieving real-time rendering speeds 100 times faster than NeRF. The concept of splatting-based rendering originated from Zwicker et al. Surface splatting~\cite{zwicker2001surface} has evolved through differentiable surface splatting for point-based geometry by Wang et al.\cite{yifan2019differentiable}, and further advancements in optimizing gradients for SH coefficients in \pre{splattings }\now{Gaussians} by Zhang et al.\cite{zhang2022differentiable}. Most recently, Kerbl et al.~\cite{kerbl20233d} have developed a method to 
\pre{simulate spatial object surfaces as anisotropic Gaussian-distributed splattings, enabling image synthesis from novel viewpoints.}
\now{simulate the surfaces of spatial objects as anisotropic Gaussians, allowing the synthesis of images from novel viewpoints.}

\section{Methodology}

\vspace{-0.0cm}

\begin{figure*}[t]
	\centering
	\includegraphics[width=1.0\linewidth]{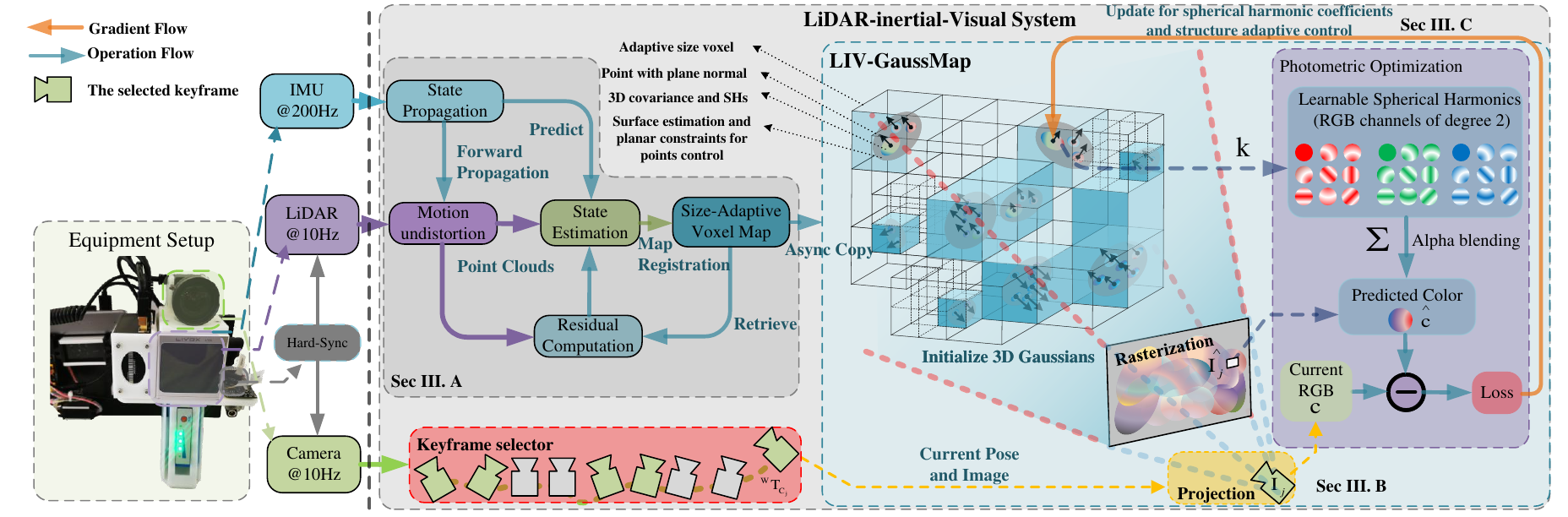}
	\caption{
\pre{The left of the image illustrates the sensory input and our equipment setup. It features an external sensor assembly comprising a synchronized LiDAR-Inertial sensor (Livox Avia) paired with a camera. On the right side, we present our algorithmic pipeline, which includes:
1. The initial representation of the scene is derived from a IESKF-based LiDAR-inertial system with size adaptive voxel, providing an initial Gaussian structure for the scene.
2. Subsequently, we optimize the Gaussians structure and spherical harmonic coefficient by photometric gradients. This involves calculating rasterization loss using images to refine the scene representation further.
}
\now{
The overview of our proposed system: The left side illustrates the sensory inputs and the configuration of the data acquisition equipment. The right side details the software pipeline, showcasing the sequence of processing steps.}
}
	\vspace{-2pt}
	\label{fig.system}
\end{figure*}
\vspace{-0.0cm}

Our system, illustrated in Fig.~\ref{fig.system}, integrates \now{with} hardware and software components. Hardware-wise, it features a hardware-synchronized LiDAR-inertial sensor paired with a camera, ensuring precise synchronization of LiDAR point clouds and image captures for accurate data alignment and fusion.

Software-wise, the process starts with LiDAR-inertial odometry~\cite{yuan2022efficient} for localization, using \pre{a size-adaptive voxel map }\now{a map with size-adaptive voxels} to represent planar surfaces. LiDAR point clouds are segmented into voxels,
\pre{where the covariance of planes is computed for initial elliptical splatting estimates (see Fig.~\ref{fig.scheme}).}
\now{where the covariance of the plane is computed for the initialization for Gaussians (Sec~\ref{sec:A}).}\pre{ The final step involves refining spherical harmonic coefficients and LiDAR Gaussian structures using images from various perspectives, leveraging photometric gradients. This approach produces a photometrically accurate LiDAR-visual map, enhancing mapping precision and visual realism. }\now{The next step involves optimizing the spherical harmonic coefficients and refining the LiDAR Gaussian structures with images captured from various perspectives using photometric gradients(Sec~\ref{sec:B} and Sec~\ref{sec:C}). This approach produces a photometrically accurate LiDAR-visual map, enhancing mapping precision and visual realism.}

\subsection{Initialization of Gaussians with LiDAR Measurement}
\label{sec:A}
Initially, we \pre{employ }\now{used} size-adaptive voxels to partition the LiDAR point cloud, drawing inspiration from the octree approach discussed in~\cite{yuan2022efficient}. 

Our adaptiveness of \now{the} voxel partition is determined \pre{based on }\now{on the basis of} evaluating a certain parameter $\eta$, which serves as an indicator to judge \pre{if }\now{whether} a voxel has a surface with planar characteristics inside.
To obtain a more precise map with a normal vector of \pre{Gaussian surface }\now{surface Gaussian}, we allow for smaller voxels and further subdivision into finer levels. If the voxel is \pre{divided small enough }\now{sufficiently divided} through multiple subdivisions, even curved surfaces can be approximated.

The voxel can be characterized by its average position $\overline{\mathbf{p}}$, the normal vector $\mathbf{n}$, and the covariance matrix $\mathbf{\Sigma_{\mathbf{n},\overline{\mathbf{p}}}}$ \pre{inside }\now{within} the voxel.

\begin{small}
\begin{align}
\overline{\mathbf{p}}=\frac{1}{N} \sum_{i=1}^{N}{ }^{w} \mathbf{p}_{i}
\end{align}
\end{small}

The covariance of \now{the} voxel $\mathbf{\Sigma_{\mathbf{n},\overline{\mathbf{p}}}}$
can be calculated as \pre{below }\now{follows}, which indicates the distribution of the points ${ }^{w} \mathbf{p}_{i}$:

\begin{small}
\begin{align}
 \mathbf{\Sigma_{\mathbf{n},\overline{\mathbf{p}}}}
 =\frac{1}{N} \sum_{i=1}^{N}\left({ }^{w} \mathbf{p}_{i}-\overline{\mathbf{p}}\right)\left({ }^{w} \mathbf{p}_{i}-\overline{\mathbf{p}}\right)^{T}
\end{align}
\end{small}

We denote the eigenvector $\mathbf{n}$, which is \pre{regarded as }\now{considered} the normal vector of the planar surface, for the covariance $\mathbf{\Sigma_{\mathbf{n},\overline{\mathbf{p}}}}$ of this hypothetical Gaussian plane~\cite{yuan2022efficient}.
The corresponding eigenvalues $\mathbf{\lambda}$ represent the distribution of this Gaussian plane in each direction. If the $\eta$, which indicates that the thickness of the planar surface is still significant, a further subdivision is performed.

\begin{small}
\begin{align}
\eta=\frac{\mathbf{\lambda}_\text{min}}{\sqrt{\mathbf{\lambda_\text{mid}}^2+\mathbf{\lambda}_\text{min}^2+\mathbf{\lambda_\text{max}}^2}}
\end{align}
\end{small}

The distribution matrix $\mathbf{\Sigma}_{\mathbf{n},\overline{\mathbf{p}}}$ is calculated to determine the approximate shape and pose of the point cloud, which contains the pose of the \pre{surface }Gaussians.
\pre{However, to seamlessly integrate these LiDAR points with surrounding points and ensure hole-free proportional scaling that upholds the integrity of the original data.}
We introduce a scaling factor $\alpha_i$ for each point, which is determined \pre{by the point density }\now{by the density of the points within the voxels}. 
\pre{This scaling factor allows for the rescaling of the points accordingly.}
\now{This scaling factor $\alpha_i$ is used to rescale the planar Gaussians.}

\begin{align}
\mathbf{\Sigma_{{ }^{w} \mathbf{p}_{i} }}=\alpha_i \mathbf{\Sigma_{\mathbf{n},\overline{\mathbf{p}}}}
\end{align}

\pre{We define the 3D radiance field of the LiDAR point cloud with the elegant form of an ellipsoidal Gaussian, represented by the following equation:}
\now{
The impact of a LiDAR point cloud on the radiance field can be determined by the following equation:
}
\begin{align}
G^\text{3D}_i(^w\mathbf{p})=
e^{-\frac{1}{2}(^w\mathbf{p}-{ }^{w} \mathbf{p}_{i})^{T} 
\mathbf{\Sigma_{{ }^{w} \mathbf{p}_{i} }}^{-1}
(^w\mathbf{p}-{ }^{w} \mathbf{p}_{i})} 
\end{align}

\subsection{Spherical Harmonic Coefficient Optimization and Map Structure Refinement with Photometric Gradients}
\label{sec:B}

\pre{The structure provided by the LiDAR-inertial system is further refined with visual photometric gradients for enhanced mapping. Further, we utilize high-order spherical harmonics, akin to those in computer graphics, for depicting view-dependent radiance surfaces.}
We utilize second-degree spherical harmonics (SHs)~\cite{cabral1987bidirectional}, which require a total of 27 harmonic coefficients for each Gaussian\pre{. And it is a balance between complexity (and thus computational cost) and accuracy. 
A more photorealistic map can be obtained by optimizing the spherical harmonic coefficients of LiDAR Gaussian through a photometric gradient. This refined map enables real-time rendering with improved interpolation and extrapolation for photorealistic mapping.
}\now{, allowing the rendering of non-Lambertian surfaces.}

The point in the world frame is $^w\mathbf{p}_i$, and the pose of the LIV system is $^w\mathbf{T}_{C_n}$.
The viewing direction for \now{the} point $^w\mathbf{p}_i$ from the pose $^w\mathbf{T}_{C_n}$ can be calculated as

\begin{small}
\begin{gather}
^{C_n}\mathbf{v}_i=\frac{^w\mathbf{T}_{C_n}^{-1} \cdot ^w\mathbf{p}_i}{\|^w\mathbf{T}_{C_n}^{-1} \cdot ^w\mathbf{p}_i\|}\\
\theta = \arccos\left(\frac{{^{C_n}\mathbf{v}_i}_z}{\sqrt{{^{C_n}\mathbf{v}_i}_x^2 + {^{C_n}\mathbf{v}_i}_y^2 + {^{C_n}\mathbf{v}_i}_z^2}}\right)\\
\phi = \arctan2({^{C_n}\mathbf{v}_i}_y,{^{C_n}\mathbf{v}_i}_x)
\end{gather}
\end{small}

The spherical harmonics function is sensitive to the viewing direction. 
\begin{small}
\begin{align}
c(\theta,\phi)=\sum_{\ell=0}^{\infty} \sum_{m=-\ell}^{\ell} k_{\ell}^{m} \sqrt{\frac{2 \ell+1}{4 \pi} \frac{(\ell-m) !}{(\ell+m) !}} P_{\ell}^{m}(\cos \theta) e^{i m \phi} 
\end{align}
\end{small}

where
$P_{\ell}^{m}(\cos \theta) e^{i m \phi} $
represents the Legendre polynomials. \now{$k_{\ell}^{m} $ is the coefficients of the spherical harmonics in each LiDAR point.}

\pre{As the Fig.\ref{fig.scheme} shows, this LIV system initially generates a dense point cloud populated with 3D Gaussians, each characterized by a position $^w\mathbf{p}_i$, a covariance matrix $\mathbf{\Sigma}_{^W\mathbf{p}_i}$, a normal vector $n$, and a color $\mathbf{c}$ and a opacity $\alpha$.}
\now{As Fig.\ref{fig.system} shows, the LIV system initially generates a dense point cloud populated with 3D Gaussians, each characterized by a position $^w\mathbf{p}_i$, a covariance matrix $\mathbf{\Sigma}_{^w\mathbf{p}_i}$, and the spherical harmonic coefficients $k_{\ell}^{m} $.}

\now{Each frame of the LiDAR-inertial sensor and the camera is synchronized through the trigger signal. }Consequently, the projection of a LiDAR point cloud from the world frame to the camera frame $C_n$ on the image plane $^{C_n}\mathbf{q}_i$ can be written as follows:

\begin{small}
\begin{align}
 ^{C_n}\mathbf{q}_i=\boldsymbol{\pi}({^w\mathbf{T}^{-1}_{C_n}} \cdot {^{w}\mathbf{p}_i})
\end{align}
\end{small}

To train a \pre{point cloud }model that predicts the image $I_{n}$. We employ the loss function as below to optimize the structure and the spherical harmonic coefficients of point clouds, that is
\begin{small}
\begin{align}
\mathcal{L}=(1-\lambda)\sum_{n=1}^{N}\sum_{\mathbf{q}\in\mathcal{R}}\left\|I_{n}(\mathbf{q})-\hat{I}_{n}(\mathbf{\mathbf{q}})\right\|+ \lambda \mathcal{L}_\textbf{D-SSIM}
\end{align}
\end{small}
where $\lambda$ is a weighting coefficient that balances the contribution of MSE and D-SSIM losses. To effectively refine the structure of the Gaussians and their spherical harmonic coefficients, we employ the Adam optimizer for the optimization of the loss function.

\now{
The novel view of the image \(\hat{I}_{j}(u)\) can be synthesized through alpha blending using the following equation:
\begin{small}
\begin{align}
\hat{I}_{n}(\mathbf{q}) = \sum_{i=1}^{M}
 [c_{i} {\sigma_i} G^\text{2D}_i(\mathbf{q}) \prod_{j=1}^{i-1}(1-\sigma_j G^\text{2D}_j(\mathbf{q}))
 ]
\label{raster}
\end{align}
\end{small}
where $G^\text{2D}_i(u)$ is the 2D Gaussian derived from $G^\text{3D}_i(x)$  through  the local affine transformation conducted in~\cite{zwicker2001surface}, $\sigma_i \in [0,1]$ is the opacity of Gaussians. $M$ is the number of Gaussians that influence the pixel.
}

\subsection{Structure Adaptive Control of 3D Gaussian map}
\label{sec:C}

\pre{The structure derived from the LiDAR-inertial system is not flawless. It may encounter difficulties in measuring surfaces made of glass or areas that have been either excessively or insufficiently scanned. }
\now{The structure derived from the LiDAR-inertial system is not flawless. It may encounter difficulties in accurately measuring surfaces made of glass or areas that have been scanned either excessively or insufficiently.}
\pre{To tackle these concerns, }
\now{To address these concerns, }we employ structure refinement to address under-reconstruction and over-dense scenarios. 

In situations where geometric features are not yet well reconstructed (under-reconstruction), noticeable positional gradients can arise within the view space. In our experiments, we establish a predefined threshold value to identify regions that require densification. We replicate the neighboring Gaussians and then employ the photometric gradient to optimize its position for structural completion.
In cases where repetitive scanning results in an overdense point cloud, we regularly evaluate \pre{its net contribution, as shown in Eq.~\eqref{net_contribe}} \now{its opacity}, and eliminate excessively non-essential regions with low opacity. This effectively reduces redundant points on the map and improves optimization efficiency.

\pre{
The net contribution \(A_{i}^{j}(u)\) of each point is determined by:
\begin{small}
\begin{align}
A_{i}^{j}(u) = \alpha_{i}^{j}(u) \prod_{k=1}^{i-1}(1-\alpha_{k}^{j}(u))
\label{net_contribe}
\end{align}
\end{small}
Where, the opacity \(\alpha_{i}^{j}(u)\) for each point is calculated using the formula:
\begin{small}
\begin{align}
\alpha_{i}^{j}(u) = \frac{1}{\sqrt{2 \pi r^{2}}} e^{-\frac{\|p_{i}^{j}-u\|^{2}}{2 r^{2}}}
\end{align}
\end{small}
For efficiency, we consider only points within a specific maximum distance from \(u\) and with significant opacity.
\subsection{Novel View Synthesis with Gaussians}
By employing rasterization~\cite{zwicker2001surface} for the synthesis of images from a Gaussian cloud generated by LiDAR, the novel view of the image \(\hat{I}_{j}(u)\) can be synthesized through alpha blending using the following equation:
\begin{small}
\begin{align}
\hat{I}_{n}(u) = \sum_{i=1}^{M} A_{i}^{j}(u) c_{i}^{j}
\end{align}
\end{small}
}

\section{Real-world experiments}

As shown in Table~\ref{tab:dataset}, the detailed device configurations for the four evaluation datasets are presented. To thoroughly evaluate the effectiveness of our algorithm, we purposely conducted tests on two publicly accessible datasets and two proprietary datasets that encompass a wide range of LiDAR modalities. Specifically, we used the FusionPortable dataset~\cite{jiao2022fusionportable}, which features repetitive scanning LiDAR, and the FAST-LIVO dataset~\cite{zheng2022fast}, which includes non-repetitive LiDAR data from public sources. 
\pre{In comparison }\now{Compared} to existing datasets, ours offers a comprehensive array of LiDAR modalities captured in both indoor and outdoor environments, ensuring robust hardware synchronization and accurate \now{calibration of} intrinsic~\cite{furgale2013unified} and extrinsic~\cite{yuan2021pixel} parameters\pre{\ calibration}. \pre{Additionally }\now{In addition}, we provide ground-truth structures in the form of point clouds to facilitate structure accuracy evaluation. For the execution of our \pre{Mapping }\now{mapping} system, we employ a high-performance desktop computer powered by an Intel Core i9 12900K 3.50GHz processor and a single NVIDIA GeForce RTX 4090.


\begin{table*}[htp!]
\centering
\caption{Specifications of LiDAR-inertial-Visual system in tested datasets}
\begin{tabular}{c|c|c|c|c|c}
\Xhline{1.2pt}
\multicolumn{2}{c|}{Dataset} & FAST-LIVO~\cite{zheng2022fast}& FusionPortable~\cite{jiao2022fusionportable}& Our Device I & Our Device II \\
\hline 
\multirow{6}{*}{LiDAR} & Device name & Livox Avia & Ouster OS1-128 & RealSense L515 & Livox Avia \\
& Points per second & 240,000 & 2,621,440 & 23,000,000 & 240,000 \\
& Scanning mechanism & Mechanical, non-repetitive & Mechanical, repetitive& Solid-state & Mechanical, non-repetitive \\
& Range & \( 3\mathrm{~m}-450 \mathrm{~m} \) & \( 1\mathrm{~m}-120 \mathrm{~m} \) & \( 9 \mathrm{~m} \) - \( 25 \mathrm{~m} \) & \( 3\mathrm{~m}-450 \mathrm{~m} \) \\
& Field of View 
&$70.4^{\circ}\times 77.2^{\circ}$  
&$45^{\circ} \times 360^{\circ}$
&$70^{\circ} \times 55^{\circ}$
&$70.4^{\circ} \times 77.2^{\circ}$\\
& IMU & BM1088 & ICM20948 & BMI085 & BMI088 \\
\hline 
\multirow{4}{*}{Camera} & Device name & MV-CA013-21UC & FILR BFS-U3-31S4C & RealSense L515 & MV-CA013-21UC \\
& Shutter mode & Global shutter & Global shutter & Rolling shutter & Global shutter \\
& Resolution & \( 1280 \times 1024 \) & \( 1024 \times 768 \) & \( 1920 \times 1080 \) & \( 1280 \times 1024 \) \\
& Field of View 
&$72^{\circ}\times 60^{\circ}$  
&$66.5^{\circ} \times 82.9^{\circ}$
&$70^{\circ} \times 43^{\circ}$
&$72^{\circ} \times 60^{\circ}$\\
\hline 
\multicolumn{2}{c|}{Synchronization} & \checkmark & \checkmark & \checkmark & \checkmark \\
\hline 
\multicolumn{2}{c|}{Grouth Truth of Structure} & \texttimes & \checkmark & \checkmark & \checkmark \\
\hline
\multicolumn{2}{c|}{Dataset Sequence} &
{\begin{minipage}{2cm}\centering HKU\_LSK(indoor)\\HKU\_MB(outdoor)\end{minipage}}
&
HKUST\_indoor
& 
UST\_RBMS
&
{\begin{minipage}{2cm}\centering UST\_C2\_indoor\\UST\_C2\_outdoor\end{minipage}}
 \\
\Xhline{1.2pt}
\end{tabular}
\label{tab:dataset}
\end{table*}

\pre{In the subsequent section, we conducted comparative and ablation experiments to evaluate the optimization factors, which revealed a considerable improvement in both PSNR and the structural score.}

\subsection{Evaluation for Novel View Synthesis with previous work}

\begin{figure}[t]
	\centering
	\includegraphics[width=1.0\linewidth]{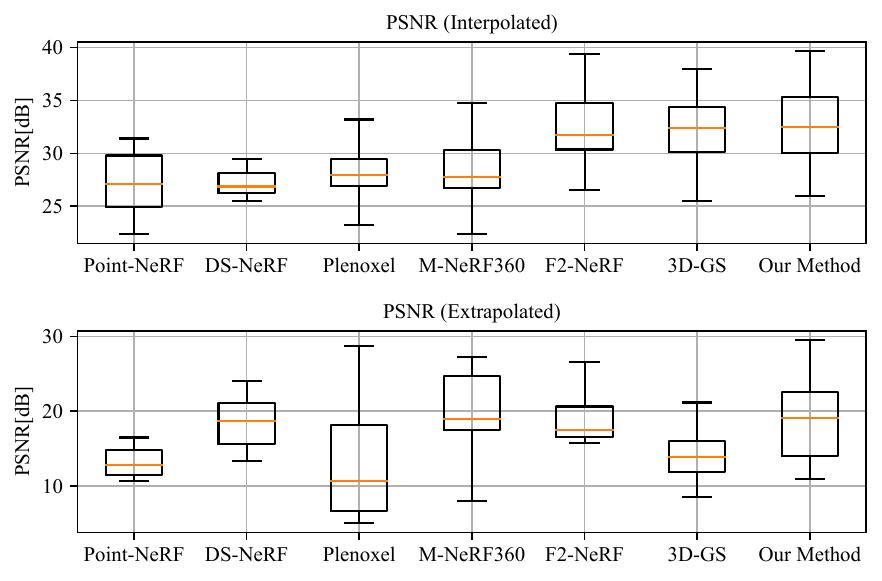}
	\caption{
This box plot illustrates the comparative performance of the leading method and our approach in terms of interpolation and extrapolation across datasets by PSNR values.
 }
	\vspace{-2pt}
	\label{fig.box}
\end{figure}

\begin{figure}[htp!]
	\centering
	\includegraphics[width=1\linewidth]{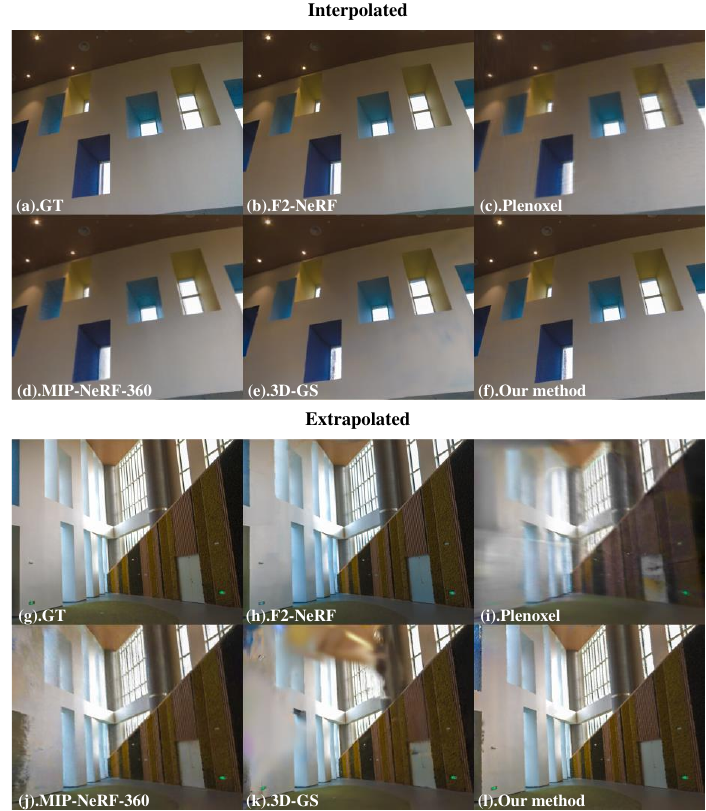}
	\caption{  
We present a comprehensive comparison between our proposed method and the state-of-the-art technique, showcasing the results of both interpolation and extrapolation for synthesizing novel viewpoints. The upper row exhibits interpolated views, while the bottom row demonstrates extrapolated viewpoint synthesis. 
 }
	\vspace{-2pt}
	\label{fig.SOTA}
\end{figure}

\pre{
As shown in Fig.~\ref{fig.SOTA}, 
we evaluated the performance of our mapping system on a real-world dataset for rendering quality against other state-of-the-art frameworks, namely Plenoxel~\cite{fridovich2022plenoxels}, F2-NeRF~\cite{xu2022point}, and 3d Gaussian splatting~\cite{kerbl20233d}. As shown in Table~\ref{tab:SOTA} and the boxplot shown in Fig. \ref{fig.box}, our framework achieved a significant improvement of 5dB in PSNR for extrapolation, indicating superior rendering quality. Additionally, we observed a competitive performance in interpolation PSNR as well.}

\now{
In our study, we evaluated the performance of our mapping system on a real-world dataset, comparing it against other state-of-the-art frameworks, such as Plenoxel~\cite{fridovich2022plenoxels}, F2-NeRF~\cite{xu2022point}, DS-NeRF~\cite{deng2022depth}, Point-NeRF~\cite{xu2022point}, and 3D Gaussian splatting~\cite{kerbl20233d}, as shown in Fig.\ref{fig.SOTA}. Note that the methods marked with an asterisk (*) in Table~\ref{tab:SOTA} were enhanced using dense LiDAR point clouds.}

\vspace{-0.0cm}

\vspace{-0.0cm}

\begin{table*}[htp!]
\centering
\caption{Quantitative evaluation of our method with previous work}
\begin{tabular}{c|c c c c c c c c}
\Xhline{1.2pt}
&
{\begin{minipage}{1.5cm}\centering PSNR[dB]$\uparrow$\\(Interpolate)\end{minipage}}
&
{\begin{minipage}{1.5cm}\centering SSIM$\uparrow$\\(Interpolate)\end{minipage}}
&
{\begin{minipage}{1.5cm}\centering LPIPS$\downarrow$\\(Interpolate)\end{minipage}}
&
{\begin{minipage}{1.5cm}\centering PSNR$\uparrow$\\(Extrapolate)\end{minipage}}
&
{\begin{minipage}{1.5cm}\centering SSIM$\uparrow$\\(Extrapolate)\end{minipage}}
&
{\begin{minipage}{1.5cm}\centering LPIPS$\downarrow$\\(Extrapolate)\end{minipage}}
&
{\begin{minipage}{1.5cm}\centering Cost  \\
Time
\end{minipage}}
&
{\begin{minipage}{1.5cm}\centering FPS \\
\end{minipage}}
\\
\hline
Point-NeRF$^{*}$\cite{xu2022point}& 27.331 & 0.872 & 0.457 &  13.117 & 0.631 & 0.610&30m20s&0.06 \\
DS-NeRF$^{*}$\cite{deng2022depth}& 27.178 & 0.831 & 0.428 & 18.534 & 0.712 & 0.533 &4h&0.06\\
3D-GS$^{*}$\cite{kerbl20233d} &  \cellcolor{yellow!35}31.900 & \cellcolor{yellow!35}0.913 &\cellcolor{yellow!35}0.241 & 15.112 & 0.647 & \cellcolor{yellow!35}0.503 & \cellcolor{orange!35}14m10s&\cellcolor{orange!35}47\\
Plenoxel\cite{fridovich2022plenoxels}& 26.744 & 0.844 & 0.452 & 12.916 & 0.628 & 0.575 &25m38s& 6.12 \\
M-NeRF360\cite{barron2022mip}& 28.446 & 0.820 & 0.444 & \cellcolor{orange!35}19.213 & \cellcolor{yellow!35}0.726 & 0.526 &8h&0.05\\
F2-NeRF\cite{wang2023f2} &  \cellcolor{orange!35}32.556 &\cellcolor{red!35}0.941 & \cellcolor{orange!35}0.193 &\cellcolor{yellow!35}19.100 & \cellcolor{orange!35}0.764 & \cellcolor{orange!35}0.387 &28m34s&13.9\\
3D-GS\cite{kerbl20233d}&  31.899 & 0.913 &0.240 & 15.111& 0.647 & 0.502 & \cellcolor{red!35}8m11s &\cellcolor{red!35}131 \\
Our method &\cellcolor{red!35}32.787 & \cellcolor{orange!35}0.926 & \cellcolor{red!35}0.190 & \cellcolor{red!35}19.220 &\cellcolor{red!35}0.803 &\cellcolor{red!35}0.331 & \cellcolor{yellow!35}14m25s&\cellcolor{yellow!35}43\\

\Xhline{1.2pt}
\end{tabular}
\label{tab:SOTA}
\end{table*}

\vspace{-0.0cm}
\vspace{-0.0cm}

\now{As depicted in Fig. \ref{fig.box},
the top row presents interpolated views, while the bottom row elucidates the synthesis of extrapolated viewpoints.}

Our algorithm demonstrates competitive performance in the peak signal-to-noise ratio (PSNR) compared to other state-of-the-art (SOTA) algorithms, such as 3DGS, in particular in the extrapolated perspective. This can be attributed to the inherent advantage of our LiDAR system, which provides relatively precise structural observations.
\now{
Furthermore, we compared our method with other latest implicit representation approaches, such as F2NeRF and MIP-NeRF360, and found that our results are still competitive. In particular, our training and rendering speeds are faster, highlighting the efficiency of our approach.
}

\subsection{Ablation Study for rendering performance with LiDAR structure}

\begin{table*}[htp!]
\centering
\renewcommand\arraystretch{1.0}
\caption{Ablation study for map structure optimization}
\begin{tabular}{p{1.8cm}<{\centering}|p{1.4cm}<{\centering}|p{2.2cm}<{\centering} p{2.2cm}<{\centering} p{2.1cm}<{\centering} p{2.1cm}<{\centering} p{1.9cm}<{\centering} p{0.9cm}<{\centering}}
\Xhline{1.2pt}
Metric & Method & HKU\_MB(outdoor) & HKU\_LSK(indoor) & UST\_C2\_outdoor & UST\_C2\_indoor & UST\_RBMS & Avg.\\
\hline
\multirow{4}{*}{\begin{minipage}{2cm}\centering PSNR[db]$\uparrow$\\(Interpolated)\end{minipage}}
&Case I&\cellcolor{orange!35}24.390 &\cellcolor{orange!35}31.222 & 31.843 & 31.721 & \cellcolor{red!35}31.663 & 30.168 \\
&Case II&24.341 & 25.964 & 31.983 & 29.625 & 30.211 & 28.425 \\
&Case III&24.240 & 31.045 &\cellcolor{orange!35}33.229 &\cellcolor{orange!35}31.975 & 31.047 &\cellcolor{orange!35}30.307 \\
&Case IV&\cellcolor{red!35}25.140 & \cellcolor{red!35}31.597 & \cellcolor{red!35}33.644 & \cellcolor{red!35}32.726 & \cellcolor{orange!35}31.277 & \cellcolor{red!35}30.877 \\
\hline

\multirow{4}{*}{\begin{minipage}{2cm}\centering SSIM$\uparrow$\\(Interpolated)\end{minipage}}
&Case I&0.793 & 0.798 & 0.897 & 0.916 & \cellcolor{red!35}0.872 & 0.856 \\
&Case II&\cellcolor{orange!35}0.814 & 0.780 & 0.895 & 0.891 & 0.864 & 0.849 \\
&Case III&0.809 & \cellcolor{orange!35}0.804 & \cellcolor{orange!35}0.909 & \cellcolor{orange!35}0.918 & 0.868 & \cellcolor{orange!35}0.862 \\
&Case IV&\cellcolor{red!35}0.825 & \cellcolor{red!35}0.805 & \cellcolor{red!35}0.916 & \cellcolor{red!35}0.926 & \cellcolor{orange!35}0.870 &\cellcolor{red!35}0.868 \\
\hline

\multirow{4}{*}{\begin{minipage}{2cm}\centering LPIPS$\downarrow$\\(Interpolated)\end{minipage}}
&Case I&0.316 & 0.277 & 0.115 & 0.219 & \cellcolor{red!35}0.338 & 0.253 \\
&Case II&0.304 & 0.292 & 0.147 & 0.219 & 0.358 & 0.264 \\
&Case III&\cellcolor{orange!35}0.301 & \cellcolor{orange!35}0.273 & \cellcolor{orange!35}0.101 & \cellcolor{orange!35}0.195 & 0.349 &\cellcolor{orange!35}0.244 \\
&Case IV&\cellcolor{red!35}0.296 & \cellcolor{red!35}0.259 & \cellcolor{red!35}0.094 & \cellcolor{red!35}0.190 & \cellcolor{orange!35}0.341 &\cellcolor{red!35}0.236 \\
\hline
\multirow{4}{*}{\begin{minipage}{2cm}\centering PSNR[db]$\uparrow$\\(Extrapolate)\end{minipage}}
&Case I&15.144 & 23.831 & 24.426 & 18.657 & 23.868 & 21.185 \\
&Case II&\cellcolor{orange!35}16.503 & 22.400 & \cellcolor{orange!35}25.653 & \cellcolor{red!35}20.511 & 23.792 & 21.772 \\
&Case III&16.178 & \cellcolor{red!35}24.821 & 25.047 & 18.964 & \cellcolor{orange!35}24.879 &\cellcolor{orange!35}21.978 \\
&Case IV&\cellcolor{red!35}16.530 & \cellcolor{orange!35}24.808 & \cellcolor{red!35}25.912 & \cellcolor{orange!35}19.220 & \cellcolor{red!35}25.545 & \cellcolor{red!35}22.403 \\
\hline
\multirow{4}{*}{\begin{minipage}{2cm}\centering SSIM$\uparrow$\\(Extrapolate)\end{minipage}}
&Case I&0.403 & 0.680 & 0.570 & 0.766 & \cellcolor{orange!35}0.849 & 0.654 \\
&Case II&0.441 & 0.657 & \cellcolor{red!35}0.674 & 0.771 & 0.847 &\cellcolor{orange!35}0.679 \\
&Case III&\cellcolor{orange!35}0.451 &\cellcolor{red!35}0.686 & 0.612 & \cellcolor{orange!35}0.775 & 0.848 & 0.675 \\
&Case IV&\cellcolor{red!35}0.470 & \cellcolor{orange!35}0.684 & \cellcolor{orange!35}0.648 & \cellcolor{red!35}0.801 & \cellcolor{red!35}0.851 &\cellcolor{red!35}0.691 \\
\hline
\multirow{4}{*}{\begin{minipage}{2cm}\centering LPIPS$\downarrow$\\(Extrapolate)\end{minipage}}
&Case I&0.530 & 0.402 & 0.307 & 0.370 & 0.389 & 0.399 \\
&Case II&0.494 & \cellcolor{orange!35}0.355 & \cellcolor{orange!35}0.302 & \cellcolor{red!35}0.314 & 0.393 &\cellcolor{orange!35}0.372 \\
&Case III&\cellcolor{orange!35}0.482 & 0.356 & 0.314 & 0.341 & \cellcolor{orange!35}0.382 & 0.375 \\
&Case IV&\cellcolor{red!35}0.479 & \cellcolor{red!35}0.348 & \cellcolor{red!35}0.275 & \cellcolor{orange!35}0.336 & \cellcolor{red!35}0.371 & \cellcolor{red!35}0.362 \\
\hline
\multirow{4}{*}{Cost time[min]$\downarrow$}
&Case I&26m9s & 16m58s & 24m16s & 14m15s & 20m43s & 20m28s \\
&Case II&34m14s & 17m3s & 25m55s & 17m38s & 24m5s & 23m47s \\
&Case III&19m37s & 14m19s & 16m26s & 12m20s & 18m54s & 16m19s \\
&Case IV&18m19s & 13m41s & 16m33s & 13m4s & 18m46s & 16m5s \\
\Xhline{1.2pt}
\end{tabular}
\label{tab:Ablation}
\end{table*}
\pre{Based on the experiments above, we have discovered that 3D-GS exhibits remarkable competitiveness. Consequently, our subsequent experiments will primarily revolve around comparing the 3D-GS technique with our LiDAR-assisted Gaussian structure construction approach.As shown in Table~\ref{tab:Ablation},}  To validate the effectiveness of our algorithm, we progressively integrated our optimized methods and monitored the corresponding changes in PSNR. We designed several comparative experiments as the following cases for ablation analysis\now{\ (Table~\ref{tab:Ablation})}.

Case I: Implemented 3D-GS as a baseline.
Case II: LiDAR initialization for Gaussians, showing map reconstruction without visual optimizations \now{for Gaussian structure}.
Case III: Enhanced Case II with photometric gradients to optimize \pre{point cloud distribution, improving map accuracy and robustness}\now{the Gaussian position}.
Case IV: Further refinement using photometric gradients for Gaussian pose optimization\pre{\ , enhancing map quality}.

\now{By} comparing Case I and Case II, the \pre{LiDAR }performance varies with \pre{scene complexity }\now{the complexity of the scene}. In complex structures like "HKU\_MB\_(outdoor)", LiDAR's accuracy decreases, potentially lowering PSNR. In simpler scenes like "UST\_C2\_outdoor", LiDAR achieves precise estimations, enhancing PSNR, especially in extrapolation tasks. In Case III, optimizing point-cloud distribution speeds up the process but may reduce PSNR.

Ultimately, our method (Case IV) enhances PSNR by refining the map structure, consistently outperforming 3D-GS across all scenes in both interpolation and extrapolation. We also evaluated our method with solid-state LiDAR (RealSense L515). Given its restricted measurement range,  we only conducted experiments in indoor scenes. The results demonstrate that our approach consistently maintains a superior level of PSNR.

\subsection{Structure Reconstruction Evaluation}

Our study presented qualitative and quantitative results, highlighting the effectiveness of using LiDAR for initial structure optimization (Fig.~\ref{fig.structure}). Quantitatively, we evaluated our approach \pre{using Chamfer Discrepancy (CD), Earth Mover Distance (EMD), and F-score }\now{using CD (Chamfer Discrepancy)~\cite{nguyen2021point}, EMD (Earth Mover Distance)~\cite{fan2017point}, and F-score~\cite{sokolova2006beyond}}(Tab~\ref{tab:Stru}), finding significant improvements in these metrics with LiDAR-based initialization. Although the use of photometric optimization for the Gaussian structure slightly reduced structural quality, the introduction of Gaussian pose refinement showed mixed results: it improved \pre{CD (Chamfer Discrepancy)~\cite{nguyen2021point} and EMD(Earth Mover Distance)~\cite{fan2017point} but negatively impacted the F-score~\cite{sokolova2006beyond} }\now{CD and EMD but negatively impacted the F-score}. Despite some trade-offs in structural integrity for better PSNR, our method overall demonstrated superior structural metrics compared to purely visual approaches.

\begin{table}[h!]
\centering
\caption{\scriptsize{ABLATION STUDY OF DIFFERENT DESIGN CHOICES ON FUSIONPORTABLE DATASET~\cite{jiao2022fusionportable}}}
\begin{tabular}{c|c c c c c}
\Xhline{1.2pt}
&
{\begin{minipage}{1.7cm}\centering CD~\cite{nguyen2021point}$\downarrow$\end{minipage}}
&
{\begin{minipage}{1.7cm}\centering EMD~\cite{fan2017point}$\downarrow$\end{minipage}}
&F-score~\cite{sokolova2006beyond}$\uparrow$
\\
\hline
Case I& 0.149 &  0.698 & 0.544 \\
Case II &0.114 & \cellcolor{orange!35}0.553 & \cellcolor{red!35}0.807 \\

Case III & \cellcolor{orange!35}0.109 & 0.614 & 0.682 \\

Case IV & \cellcolor{red!35}0.107 & \cellcolor{red!35}0.435 & \cellcolor{orange!35}0.751 \\

\Xhline{1.2pt}
\end{tabular}
\label{tab:Stru}
\end{table}

\newpage
\section{Conclusion}

\pre{We propose LiDAR-Inertial-Visual fused real-time 3D radiance field mapping system that capitalizes on the fusion of LiDAR-inertial visual multi-modal sensors.
}
\now{We propose a 3D radiance field mapping system that capitalizes on the fusion of LiDAR-inertial visual multimodal sensors, offering high-quality and real-time rendering capabilities.}

\pre{
Our approach leverages the precise surface measurement capabilities of LiDAR, coupled with the adaptive voxel feature innate to the LiDAR-inertial system, facilitating rapid initial scene structure acquisition. However, it's inevitable to encounter critical injection angles during LiDAR scanning, which can lead to unreasonably distributed or inaccurately measured point clouds.
}
\now{
Our method utilizes the accurate surface measurement features of LiDAR, along with the inherent adaptive voxel characteristic of the LiDAR-inertial system, to enable quick acquisition of the initial scene structure.
Additionally, we utilized visual observations to further optimize the LiDAR structure by photometric gradient, enhancing PSNR performance. 
Even though the training and rendering speeds are slower than the 3D-GS due to the dense LiDAR points, our method still shows real-time rendering performance.
}

\pre{
Through extensive real-world experiments, we have consistently demonstrated that our algorithm achieves superior geometric structures. Moreover, it produces novel view images with higher PSNR than other state-of-the-art visual-based methods, both for extrapolated and interpolated poses.
}
\now{
Through extensive real-world experiments, our approach has proven its ability to achieve superior geometric structures and high rendering quality, both extrapolated and interpolated, compared to other leading visual-based techniques.
}

\appendices

\ifCLASSOPTIONcaptionsoff
  \newpage
\fi



%

\bibliographystyle{IEEEtran}
\bibliography{root}







\end{document}